\documentclass[runningheads]{llncs}
\usepackage[utf8]{inputenc}
\usepackage{graphicx}
\usepackage{amsmath,amssymb} 
\usepackage{color}
\usepackage{subfigure}
\usepackage[width=122mm,left=12mm,paperwidth=146mm,height=193mm,top=12mm,paperheight=217mm]{geometry}
\begin{document}
\pagestyle{headings}
\mainmatter
\def\ECCV18SubNumber{1}  
\newcommand*\samethanks[1][\value{footnote}]{\footnotemark[#1]}

\title{Learning Sampling Policies for\\ Domain Adaptation}
\titlerunning{Learning Sampling Policies for Domain Adaptation}
\authorrunning{Patel, Y., and Chitta, K., and Jasani, B.}
\author{Yash Patel\thanks{Equal Contribution} \and Kashyap Chitta\samethanks \and Bhavan Jasani\samethanks \\
\institute{The Robotics Institute, Carnegie Mellon University\\
\email{\{yashp, kchitta, bjasani\}@andrew.cmu.edu}}}

\maketitle

\begin{abstract}
We address the problem of semi-supervised domain
adaptation of classification algorithms through deep Q-learning. The core idea is to consider the predictions of a source domain network on target domain data as noisy labels, and learn a policy to sample from this data so as to maximize classification accuracy on a small annotated reward partition of the target domain. Our experiments show that learned sampling policies construct labeled sets that improve accuracies of visual classifiers over baselines. 

\end{abstract}

\section{Introduction}
\label{sec:introduction}

Dataset bias \cite{torralba2011unbiased} is a well-known drawback of supervised approaches to visual recognition tasks. In general, the success of supervised recognition models, both of the traditional and deep learning varieties, is restricted to data from the domain it was trained on \cite{tzeng2017adversarial}.
The common approach to handle this is fairly straightforward: pre-trained deep models perform well on new domains when they are \textit{fine-tuned} with a sufficient amount of data from the new distribution. However, data for fine-tuning needs to be annotated. In many situations, labeling enough data for this approach to be effective is still prohibitively expensive.

Recent work on domain adaptation address this problem by aligning the features extracted from the network across the source and target domains, without any labeled target samples. The alignment typically involves minimizing some measure of distance between the source and target feature distributions, such as correlation distance \cite{sun2016coral}, maximum mean discrepancy \cite{long2015transferable},  or adversarial discriminator accuracy \cite{ganin2015backprop,tzeng2015simultaneous,tzeng2017adversarial}. 


In this work, we explore the semi-supervised domain adaptation problem. We assume that we can collect data in the target domain, as well as annotate a small fraction of it, and we have a fixed \textit{budget} for annotation. This has been extensively studied under the field of active learning \cite{cohn1994active,settles10active}, where the goal is to obtain better predictive models than those trained on equal amounts of i.i.d.  data by deciding which examples to annotate from a large unlabeled dataset. However, active learning methods are inherently designed for a target domain directly, and do not make use of the extensive amount of annotated data we have in the source domain.

We propose a reinforcement learning based formulation of the semi-supervised domain adaptation problem. In active learning, we need to choose a subset of the data to annotate and train from. We hypothesize that we could better use our annotation budget if we label a 'reward partition', used to generate rewards for a deep Q-network. Knowledge from the source domain could be coupled with this Q-agent to potentially give us a large quantity of well-labeled data in the target domain, which could not be achieved independently through unsupervised domain adaptation or active learning.

Inspired by a similar approach for action recognition \cite{yeung2017learning}, we aim to use our Q-network to learn a policy for sampling from noisily labeled data in the target domain. A classifier trained on the source domain is used to generate these noisy annotations for the entire target dataset. The agent is rewarded for sampling data from the target domain, that when used to train a new classifier, leads to high accuracies on the annotated reward partition. 

We evaluate our approach on the Office-31 dataset, a widely accepted benchmark for testing real-world visual domain adaption methods \cite{saenko2010adapting}, comparing our learned policies to baselines, and state-of-the-art unsupervised domain adaptation methods.

\section{Method}
\label{sec:method}
\begin{figure*}[t!]
\centering
\includegraphics[width=\textwidth]{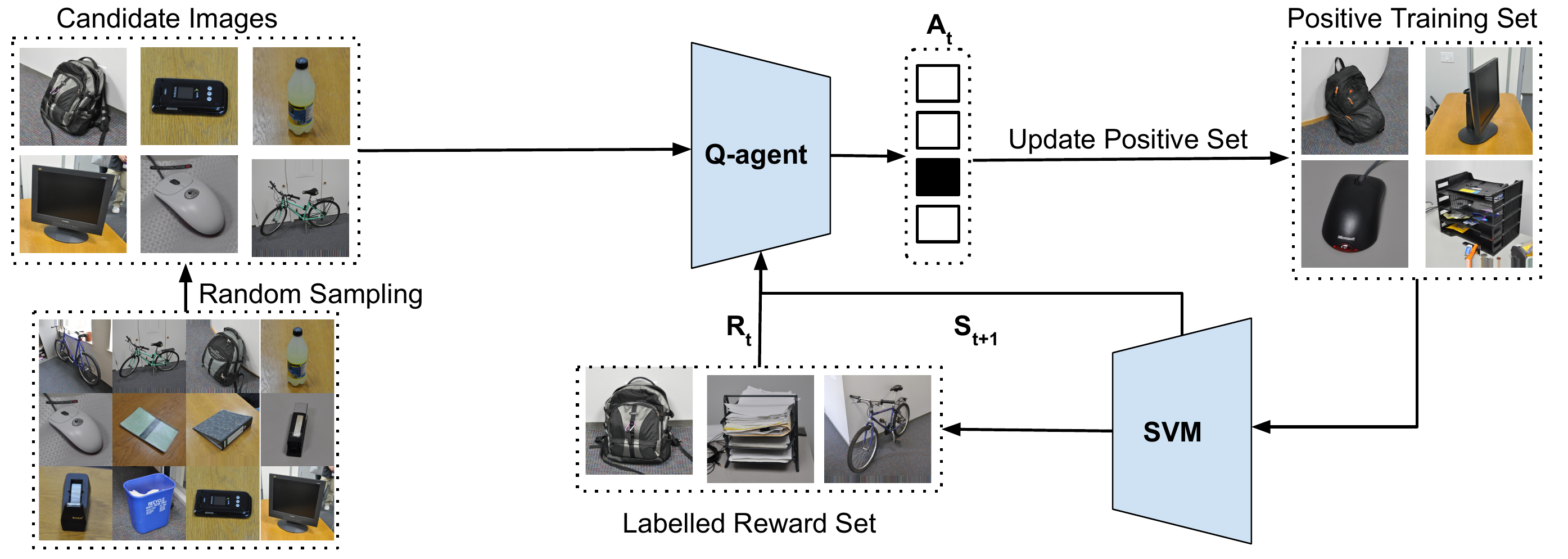}
\caption{Overall Method}
\label{fig:overall_method}
\end{figure*}

In this section, we describe the proposed method for semi-supervised domain adaptation for a $n$-way classification problem. The training data consists of images from two different domains, we will refer to these domains as $D_{s}= \{(x_s^i, y_s^i)\}_{i=1}^{N_s}$ (source domain) and $D_{t}= \{(x_t^i)\}_{i=1}^{N_t}$ (target domain).
An overview of entire method is shown in Fig \ref{fig:overall_method}. It consists of the following components:
\begin{itemize}
\item Deep convolutional neural network based \textbf{source classifier}, trained for classification of the source domain $D_{s}$ images into $n$ object categories.
\item Binary Support Vector Machine (SVM) used as a \textbf{domain discriminator}, to help select a held out subset of target domain samples $S_{rew}$ for generating rewards, and initialize a training set $S_{pos}$ for the multi-class SVM.
\item Multi-class SVM based \textbf{target classifier}, for classification of the target domain $D_{t}$ images into $n$ object categories.
\item Deep \textbf{Q-agent} sampling an image from the target domain $D_{t}$ every iteration, to be added to $S_{pos}$.

\end{itemize}

\noindent
\textbf{Source Classifier.}
For image feature representations, we make use of a ResNet-50 \cite{he2016deep} architecture pretrained on ImageNet \cite{deng2009imagenet}. We choose this model for comparison to previous work \cite{kang2018attention}. In order to obtain fine-grained representations, we first fine-tune the network on source domain $D_{s}$ in a supervised setting. We denote this source classifier as $C_{src}$.

\noindent
\textbf{Domain Discriminator.}
We make a reward set $S_{rew}$, that consists of sampled target domain images that is used to evaluate the performance of classifier and in-turn used for computing rewards for the Q-agent.
In order to pick an appropriate set of images for $S_{rew}$, we train a binary SVM on image representations to classify the images as source domain or target domain. We denote this classifier as $C_{dom}$. The reward set $S_{rew}$ is then stochastically sampled from target domain based on the sample distances from the separating hyperplane of $C_{dom}$, with the sampling weight of each sample $w_{i}\propto d_{i}.$ 
The idea behind this initialization is that the samples further away from the domain classification hyperplane are more confusing and different from source domain samples and thus make a good evaluation set \cite{rai2010active}. Note that we use our budget for ground truth annotations to get the true labels for images in $S_{rew}$.

\noindent
\textbf{Target Classifier.}
The $n$-way classifier $C_{tar}$ is trained on a subset of target domain images governed by the Q-agent, which we represent as $S_{pos}$. The target classification labels are set to the predictions from source domain classifier $C_{src}$ on the images in $S_{pos}$. At each iteration of training, the action taken by the Q-agent involves updating $S_{pos}$, and this updated training set is used to train $C_{tar}$ again. For our setup, we make use of multi-class SVM for $C_{tar}$ since the number of images in target domain $D_{t}$ is limited, and we need relatively quick convergence since it is repeatedly retrained every episode. 

For the initialization of $S_{pos}$, we use $C_{dom}$ again, but now sample $l$ data points from $D_t$ with weights directly proportional to the sample distances from the classification hyperplane $w_{i}\propto \frac{1}{d_{i}}$. This follows the idea that the samples confusing the domain classifier are quite similar visually to those in the source domain, making the source classifier's predictions on these more reliable.

\subsection{Q-agent}

The objective of the Q-agent is to predict the appropriate set of samples which maximize the performance of target domain classifier $C_{tar}$ on the reward set $S_{rew}$. At any given timestep $t$, the Q-agent observes the current state $s_t$, selects an action $a_t$ from its discrete action space, and receives a reward $r_t$ as well as a next state observation $s_{t+1}$. 

\noindent
\textbf{Actions.} To have a fixed length action space for every iteration, we randomly sample $n_{cand}$ samples in $D_{t}$, with replacement, giving new set of examples each iteration we call the candidate set $S_{cand}$. The Q-agent then chooses one of $n_{cand}+1$ actions, which are either to:
\begin{itemize}
\item pick a sample from $S_{cand}$ to be added to $S_{pos}$, in which case that sample is flagged and no longer chosen for $S_{cand}$ for the rest of the episode, or
\item pick none of the samples in the current $S_{cand}$ and move to the next iteration.
\end{itemize}


\noindent
\textbf{State Representation.} We formulate the states as a concatenation of two vectors, one dependent on the positive set and the other on the current candidate set. 
To get the first part of the state representation which shows the current distribution of examples in the positive set, we use a histogram of classifier confidences, obtained by using $C_{tar}$ on the current $S_{pos}$, 
We threshold the distribution over classes output by $C_{tar}$ into discrete bins of equal width. This component of the state representation has a dimensionality of $ n \times n_{bin}$ (where $n$ is the number of classes).

The second part of the state representation gives the agent a concise summary of the choices it has: for every image in $S_{cand}$, we obtain the distribution of classifier confidences for each class as a confidence vector for that sample. The concatenation of all these vectors, each of which corresponds to a specific action that the Q-agent can take, has a dimensionality of $ n \times n_{cand}$.

The entire state vector, which is the input to the Q-network, is a flattened concatenation of these two parts.

\noindent
\textbf{Rewards.} The reward for the agent at any time stamp $r_{t}$ is determined by the relative change in performance of target domain classifier $C_{tar}$ from the previous time stamp. 

\noindent
\textbf{Q-Network.} In the standard formulation of Q-learning, we use a function approximator with some weights $w$ to estimate the $Q$-value of a state-action pair $(s,a)$. 
We use a Dueling Deep Q-Network (DDQN) \cite{wang2015dueling} as our function approximator. 
We define the advantage function relating the value and Q functions:
\begin{equation}
A_\pi(s, a) = Q_\pi(s, a) - V_\pi(s)
\end{equation}
We implement the DDQN with two sequences (or streams) of fully connected layers. 
We subtract the average output over all actions from the advantage stream:

\begin{equation}
Q(s, a; \alpha,\beta) = V (s; \beta) +  A(s, a; \alpha)  - \frac{1}{|A|} \sum_{a'}A(s, a'; \alpha)
\end{equation}

where $\alpha$ are the parameters of the advantage stream, $\beta$ are the parameters of the value stream, and the weights $w$ we intend to optimize is the combination of these two sets of parameters.

\noindent
\textbf{Stabilization.} We further stabilize the training by maintaining two Q-networks \cite{van2016deep}: an online network with weights $w$, and a target network with weights $w^-$. The weights $w$ are updated using the following gradients:

\begin{equation}
	w := w + 
		\eta \left( 
			r + \gamma \max _{a' \in A} Q _{w ^-} (s', a') 
				- Q _w(s, a) 
		\right) \nabla _w 
			Q _w (s, a) .
\end{equation}

Where $Q _{w ^-}$ is generated by the target network, and $\eta$ is our learning rate. Every few iterations, the target network weights are set to be equal to the online Q-network, and are kept frozen until the next such assignment. 

We additionally found that adding weight decay as a regularizer to the objective and using a sigmoid activation on the final predicted Q values significantly improved the optimization. Since the Q values are expectations of cumulative rewards under an optimal policy, the sigmoid activation bounds these values to the range [0,1]. This is a valid restriction based on our reward structure, as the maximum possible reward of an optimal policy is 1.

\section{Experiments}
\label{sec:experiments}

We evaluate our algorithm on Office-31 dataset \cite{saenko2010adapting}. It contains images from 3 domains, \textbf{A}mazon, \textbf{W}ebcam and \textbf{D}SLR. 
Within each domain, images belong to 31 classes of everyday objects, with a fairly even class distribution. 
The dataset is imbalanced across domains, with 2,817 images in \textbf{A}, 795 images in \textbf{W}, and 498 images in \textbf{D}. With 3 domains, we have 6 possible transfer tasks, which address various forms of domain shift including resolution, lighting, viewpoint, background and dataset size.

\subsection{Experimental Setup}
The standard protocol for evaluation in unsupervised adaptation involves using all images in the source domain (with labels) and target domain (without labels) for training, and reporting performances on the entire target domain. We compare our results to other work based on the same feature extraction backbone (ResNet-50), by evaluating the final learned policy with the entire target domain dataset. The key difference between existing approaches and ours is that we use $k=3$ labels per class from the target domain in addition to the remaining unlabeled data during training. 

\subsection{Implementation Details}
\textbf{Source Classifier.} For each transfer task, we initially fine-tune our backbone network on the source domain for 30,000 iterations. We use SGD with an exponentially decaying learning rate starting at 0.003 and a batch size of 16. The 2048-dimensional pool5 feature vectors are used as representations for the images by the other components of our algorithm.

\noindent
\textbf{SVMs.} For the domain discriminator $C_{dom}$, we train a binary SVM with a linear kernel using representations from the entire source domain as one class and target domain as the other. The target classifier $C_{tar}$ is a multi-class linear kernel SVM trained in a one-vs-all manner every iteration. Both of these are implemented with the default parameters using the liblinear library \cite{fan2008liblinear}. We initialize $S_{pos}$ with $l=100$ samples for all 6 transfer tasks.

\noindent
\textbf{Q-agent.} For our state representation, we set $n_{bin}$ of the histogram to 10 and $n_{cand}$ to 20. This leads a state space of dimensionality 930 and action space of size 21. For the value stream of our DDQN, we use a single hidden layer with as many units as the dimensionality of the state. 
The advantage stream consists of two hidden layers of 512 units each. 

We use the Adam optimizer \cite{kingma2014adam} with a learning rate of 0.001. We use an $\epsilon$-greedy linear exploration policy,
reducing the value of $\epsilon$ from 1 to 0 over the 2,000 iterations of training, and train for a total of 20,000 iterations for each transfer task. We assign the weights of our online network to the target network every 10 iterations. Our models are implemented on Keras with a TensorFlow backend \cite{abadi2016tensorflow}.


\subsection{Results}
\label{sec:results_results}
We report our results in Table \ref{t:results}. As a baseline, we use the classifier trained on source domain and evaluate on target domain (without any domain adaptation) for all 6 transfer tasks. We additionally compare to the best reported results for unsupervised domain adaptation on this dataset \cite{kang2018attention}.

We observe that the learned policies do better than baselines, but fail to close the gap towards existing state-of-the-art unsupervised adaptation methods. 

\begin{table}
\centering
\caption{Comparison with state of the art}
\begin{tabular}{c | c | c | c | c | c | c}
\hline
\textbf{Method} & A $\Rightarrow$ D & A $\Rightarrow$ W & D $\Rightarrow$ A & D $\Rightarrow$ W & W $\Rightarrow$ A & W $\Rightarrow$ D \\ \hline
Baseline (ours) & $76.9\%$ & $71.4\%$ & $58.6\%$ & $91.7\%$ & $57.4\%$ & $96.8\%$ \\ 
Ganin et al. \cite{ganin2015backprop} & $79.7\%$ & $82.0\%$ & $68.2\%$ & $96.9\%$ & $67.4\%$ & $99.1\%$\\ 
Kang et al. \cite{kang2018attention} & $88.8\%$ & $86.8\%$ & $74.3\%$ & $99.3\%$ & $73.9\%$ & $100\%$\\ \hline
Ours (semi-supervised) &  $86.1\%$ &  $83.5\%$ & $64.6\%$ & $93.1\%$ & $60.8\%$ & $98.2\%$ \\
\hline
\end{tabular}
\label{t:results}
\end{table}

\section{Conclusion}
\label{sec:conclusion}
We present a reinforcement learning based approach to learn sampling policies for the purpose of domain adaptation. Our method learns to select samples for training from the target domain that maximize performance on a reward set, and in-turn improve the overall classification accuracy in the target domain. 

Unlike the existing state-of-the-art methods, we make use of fixed representations for both the source and target domain samples which hurts the performance of our method. In order to fix this, we plan to work on the integration of a \textit{labeler} into the Q-agent which will be jointly optimized with the current \textit{sampler} to tune better representations and less noisy labels for target domain. Another idea is to learn sampling policies using the representations obtained after performing unsupervised domain adaptation through existing feature alignment techniques.  

{\small
\bibliographystyle{splncs}
\bibliography{eccv2018submission}}
\end{document}